\documentclass[runningheads]{llncs}
\usepackage[T1]{fontenc}
\usepackage{graphicx}

\usepackage{acronym}
\usepackage[acronym]{glossaries}
\usepackage{amsmath, bm}
\usepackage{amsfonts}
\usepackage{xcolor}
\usepackage{xspace}
\usepackage[sort]{cite}
\usepackage[group-separator={\;}]{siunitx}
\usepackage{booktabs}
\usepackage{multirow}
\usepackage{microtype}

\usepackage[breaklinks,colorlinks,linkcolor=black,citecolor=blue,urlcolor=blue]{hyperref}
\usepackage[nameinlink,sort&compress,capitalize]{cleveref}
\Crefname{section}{Section}{Sections}
\Crefname{table}{Table}{Tables}
\crefname{table}{Table}{Tables}
\crefname{figure}{Figure}{Figures}

\newacronym{loe}{LOE}{Layout Object Encoder}
\newacronym{ocr}{OCR}{Optical Character Recognition}
\newacronym{dla}{DLA}{Document Layout Analysis}
\newacronym{mlp}{MLP}{Multi-layer Perceptron}

\makeatletter
\DeclareRobustCommand\onedot{\futurelet\@let@token\@onedot}
\def\@onedot{\ifx\@let@token.\else.\null\fi\xspace}

\def\eg{e.g\onedot}

\def\etc{\etc\onedot} 
\def\wrt{w.r.t\onedot} 
 
\def\etal{et al\onedot}
\makeatother

\newcommand{\ourwork}{BBLP\xspace}
\newcommand{\ourworklong}{Bounding Box Label Propagation\xspace}

\newcommand{\dfourla}{D$^4$LA\xspace}

\newcommand{\textBF}[1]{%
    \pdfliteral direct {2 Tr 0.3 w} 
     #1%
    \pdfliteral direct {0 Tr 0 w}
}

\begin{document}
\title{Bounding Box Label Propagation\\ for Re-Annotation of Document\\ Layout Analysis Datasets}
%
\titlerunning{Bounding Box Label Propagation}
\sisetup{group-separator={\,}} 
%
%
\author{Nick Jochum\textsuperscript{*}\inst{1}\and
Tobias Alt-Veit\inst{1} \and
Christian Schön\inst{1} \and\\
Alexander Lück\inst{1} \and
René Schuster\inst{2,3} \and
Didier Stricker\inst{2,3}}
%
\authorrunning{N.~Jochum et al.}
%
\institute{Insiders Technologies GmbH, Kaiserslautern, Germany \\
\email{\{n.jochum, t.alt-veit, c.schoen, a.lueck\}@insiders-technologies.de} \and
DFKI -- German Research Center for Artificial Intelligence \\
\email{\{rene.schuster, didier.stricker\}@dfki.de} \and
RPTU -- University Kaiserslautern-Landau, Germany
\begin{center}
    \textsuperscript{*} corresponding author
\end{center}
}

\maketitle              

\begin{abstract}
Datasets in practical document processing scenarios typically grow over time, and their class annotations undergo continuous refinement. This creates significant re-annotation efforts, which are time-consuming and costly. A promising remedy is to re-annotate only a small subset of available documents manually and apply semi-supervised learning techniques that leverage both labelled and unlabelled data. Although there are numerous approaches to tackle this problem for classification, there exists no adaptation for the problem of re-classifying object detection instances, \eg for document layout analysis. To this end, we propose \ourworklong (\ourwork), a pseudo-labelling framework for object detection. An object encoder integrates visual, textual, and positional embeddings from object detection samples to come up with a joint embedding that can be used for Label Propagation on partially annotated datasets in a plug-and-play fashion. Evaluation results indicate that the proposed approach produces high-quality class annotations of bounding boxes. In the \dfourla layout analysis dataset, it achieves a mAP of $54.0\%$, corresponding to $81.6\%$ of fully supervised performance, while using only $10\%$ labelled data. Our work demonstrates the potential of Label Propagation for object detection and lays the groundwork for reducing manual annotation efforts in real-world document processing applications.
\end{abstract}

\section{Introduction} \label{sec:intro}
The task of \gls{dla} is a fundamental step in automatic document processing, as it provides a comprehensive understanding of the physical layout of documents~\cite{binmakhashen19_dlasurvey}. This serves as an important prerequisite for downstream tasks such as table analysis~\cite{nassar22_tableformer} and key information extraction~\cite{huang22_layoutlmv3}, as well as the overarching goal of full document understanding~\cite{tang23_udop}.

Modern layout analysis approaches address this task by training object detection architectures based on CNNs~\cite{girshick14_rcnn,ren15_fasterrcnn}, 
vision transformers~\cite{dosovitskiy21_vit}, 
or vision-language models~\cite{xu20_layoutlm,li22_dit,wang24_dlaformer}
on large annotated document datasets~\cite{li20_docbank,zhong19_publaynet}. 

However, existing research datasets are too static for industrial use case demands. In fact, the required datasets in practice grow continuously and undergo repeated refinements \wrt the object classes~\cite{gamelli24_dladatasetsurvey}, while the object bounding boxes are often retained. Such refinements are demanding: Splitting a single class into two more refined subclasses already requires a full re-annotation pass through the dataset. These novel classes are often introduced as a refinement of already existing base classes and maintain their bounding box.

Annotation effort can be reduced by re-annotating only a small fraction of the dataset and applying semi-supervised learning~\cite{engelen19_sslsurvey} to leverage both the re-annotated and the remaining data. However, standard approaches for semi-supervised object detection~\cite{banerjee24_semidocseg, sohn20_stac} do not exploit bounding box information on the unlabelled samples, as they assume such information is unavailable, thereby discarding potentially valuable information.

We on the other hand, approach the problem as a transductive classification problem and employ Label Propagation~\cite{zhu02_labelprop,zhou03_localconsistency} for the re-classification of bounding boxes. This method iteratively propagates class labels from a small labelled sample set to a larger unlabelled one. Our proposed architecture allows using Label Propagation in a plug-and-play fashion for the task of re-classifying object detection labels.

\subsubsection{Our Contribution.}
We introduce \ourworklong (\ourwork), a compact framework to adapt Label Propagation to class pseudo-labelling for document layout objects from a limited amount of labelled data. 

Therein, we leverage visual, textual, and positional embeddings to come up with a joint multimodal representation of document layout objects. This representation is trained on a generic, large document layout dataset through a surrogate classification problem. As a result, \ourwork creates suitable embedding vectors that can be used for Label Propagation on so-far unseen and only partially labelled datasets, even with different class definitions.

We evaluate the performance of \ourwork on several large-scale public document datasets. Specifically, we assess the quality of the automatically generated bounding box class annotations using two evaluation protocols: The first one compares the generated pseudo-labels directly with the corresponding ground-truth annotations, while the second one analyses the performance of an object detector trained on these noisy pseudo-labels. Moreover, we demonstrate the importance of each included modality through an extensive ablation study. Our results indicate that \ourwork can recover the majority of supervised training performance from only a limited set of labelled data.

Thus, the presented \ourwork framework provides a straightforward option to re-annotate large-scale datasets with significantly reduced manual effort.

\section{Related Work}
Object detection is concerned with simultaneously localizing and classifying objects in images. Early deep learning methods employed a two-stage approach, first proposing candidate regions and then classifying them and fitting a bounding box~\cite{girshick14_rcnn,ren15_fasterrcnn}. Single-stage detectors later emerged as a computationally efficient alternative, directly predicting bounding boxes and class labels in a unified network~\cite{redmon16_yolo,lin20_focalloss}. 
More recently, transformer-based architectures have achieved state-of-the-art performance by eliminating rule-based design choices such as anchor boxes and non-maximum suppression~\cite{carion20_detr,zhang23_dino}.

\acrlong{dla} can be viewed as an instance of object detection, where the goal is to identify and classify layout elements such as text blocks, figures, and tables within document images~\cite{binmakhashen19_dlasurvey}. Traditional approaches relied on rule-based heuristics
, but modern approaches usually fine-tune object detection transformers for the downstream layout analysis task. The LayoutLM family introduced multimodal pre-training that jointly models text, layout position, and visual features~\cite{xu20_layoutlm}. Similarly, the Document Image Transformer~(DiT)~\cite{li22_dit} employs self-supervised pre-training on large document corpora. 

While large-scale benchmarks for document layout analysis~\cite{zhong19_publaynet,pfitzmann22_doclaynet} have enabled significant progress, they most often do not capture the dynamics of practical scenarios. In real-world document processing, datasets grow incrementally, and class taxonomies evolve as new document types are encountered, or finer distinctions are required~\cite{gamelli24_dladatasetsurvey}. Each change in class ontology requires a complete re-annotation pass through the dataset, which comes with high manual efforts. This motivates semi-supervised approaches that can propagate annotations from a small manually labelled subset to the remaining data.

Semi-supervised learning has a rich history in classification tasks~\cite{chapelle06_ssl,engelen19_sslsurvey}, with pseudo-labelling~\cite{rizve21_ups} and consistency regularization~\cite{tarvainen17_meanteacher} emerging as dominant paradigms. Extending these ideas to object detection presents additional challenges, as both localization and classification must be addressed simultaneously.

Contemporary semi-supervised object detection methods predominantly adopt a teacher-student framework. STAC~\cite{sohn20_stac} pioneered this direction with an offline pseudo-labelling approach that generates complete bounding box annotations on unlabelled images. Unbiased Teacher~\cite{liu21_unbiasedteacher} introduces an online exponential moving average framework with focal loss~\cite{lin20_focalloss} to mitigate class imbalance. Soft Teacher~\cite{xu21_softteacher} further refines this approach by weighting classification losses according to prediction confidence and employing box jittering to improve localization quality. Subsequent works such as Humble Teacher~\cite{tang21_humbleteacher}, Dense Teacher~\cite{zhou22_denseteacher}, and PseCo~\cite{li22_pseco} propose various enhancements including soft pseudo-labels, dense prediction maps, and explicit mechanisms for localization precision.

Recent efforts have begun exploring semi-supervised learning specifically for document layout analysis. Kallempudi~\etal~\cite{kallempudi22_ssoddla} apply it to graphical object detection in documents, while SemiDocSeg~\cite{banerjee24_semidocseg} combines pseudo-labelling with domain adaptation for layout segmentation.

Notably, all the aforementioned semi-supervised methods generate pseudo-labels for both class assignments and bounding box coordinates. This design choice is appropriate when no prior annotations exist, but introduces unnecessary complexity when bounding boxes are already available. Our setting fundamentally differs: we assume that bounding box annotations already exist, but have outdated or incomplete class labels. This scenario arises naturally when class taxonomies evolve due to changing requirements. Rather than generating complete pseudo-boxes, \ourwork focuses exclusively on re-classifying existing bounding boxes without modifying their spatial coordinates. This decomposition allows us to leverage Label Propagation, a method specifically designed for classification problems.

Label Propagation is a graph-based semi-supervised learning method that iteratively spreads labels from annotated to unannotated samples based on their similarity. The foundational algorithm was introduced by Zhu and Ghahramani~\cite{zhu02_labelprop}, who formulated Label Propagation as computing harmonic functions on a similarity graph. Concurrently, Zhou~\etal~\cite{zhou03_localconsistency} proposed a related method termed label spreading, which incorporates a normalized graph Laplacian and allows soft label assignments. 

The key insight underlying these methods is the smoothness assumption: Samples that are close in feature space are likely to share the same label. By constructing a graph where edge weights reflect pairwise similarities, labels can be propagated along paths of high similarity even when direct neighbours are unlabelled. This transductive approach leverages the global structure of the data distribution rather than relying solely on local decision boundaries.

Label Propagation has seen renewed interest in the deep learning era. Iscen~\etal~\cite{iscen21_labelprop} demonstrated that propagating labels in the embedding space of neural networks yields competitive semi-supervised classification results. More recently, Huang~\etal~\cite{huang21_correctsmooth} showed that combining Label Propagation with simple base predictors can match or exceed graph neural network performance, highlighting the continued relevance of these classical methods.

We build upon this foundation by adapting Label Propagation to the object detection setting. Rather than operating on image-level features, \ourwork constructs embeddings for individual bounding box instances by integrating visual, textual, and positional information. This multimodal representation enables effective Label Propagation across layout objects, allowing class labels to spread from a small annotated subset to the full dataset while preserving the original bounding box coordinates.

\section{\ourworklong} \label{sec:method}

\begin{figure}[t]
    \centering
    \includegraphics[width=1\linewidth]{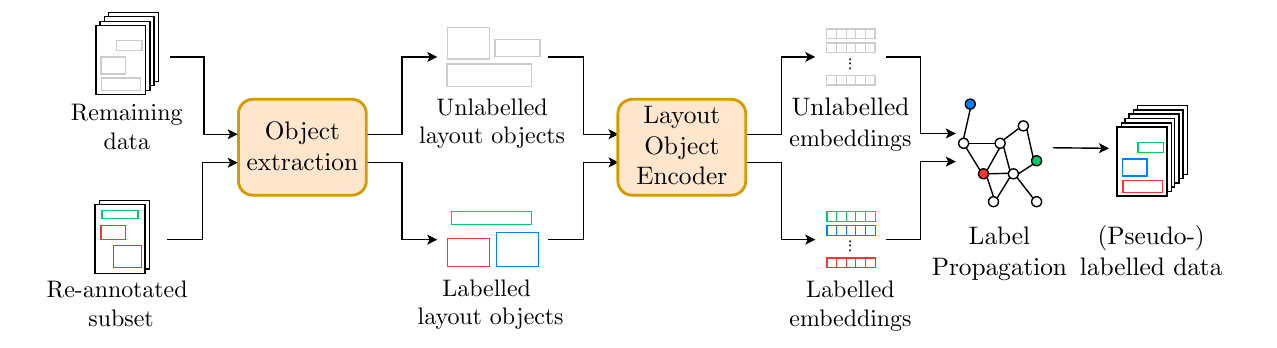}
    \caption{\ourwork framework overview. For all extracted object regions, the proposed multi-modal \acrlong{loe} embeds all layout object into a common vector space. Label Propagation then exploits this representation and generates pseudo-labels for unlabelled objects by transferring label information from the labelled objects.}
    \label{fig:loe_overview}
\end{figure}

Our pseudo-labelling framework consists of three core components. Given the complete set of all bounding boxes in a dataset and their corresponding document images, we first create several embeddings tailored specifically towards the \gls{dla} task. These embeddings take visual and textual features of individual bounding boxes as well as their position, size, and surrounding context into account. Based on those individual layout object embeddings, we then come up with a uniform embedding that combines all relevant information in a single latent space. Using this uniform embedding, we finally perform automatic Label Propagation to re-label all objects in the given dataset, based on labels provided for a small subset of objects. \Cref{fig:loe_overview} provides an overview of the proposed framework.

\subsection{Input Modalities}
Common \gls{dla} tasks include a multitude of different classes, which usually require different modalities to capture all relevant information for classification. Images or diagrams often rely on visual features to correctly locate and classify them in the context of layout analysis. Text-based classes, such as paragraphs or footnotes, are better distinguished by textual features. 
Headings or image captions in contrast often require positional context, such as their locations within the document or relative positions of neighbouring objects, for accurate distinction.
To correctly capture all relevant multi-modal information, we propose a set of embedding functions tailored specifically to the problem of bounding box re-annotation.

\subsubsection{Visual Embedding.}
\ourwork leverages a visual embedding model to encode the visual appearance of individual layout objects. Many of those embedding models have the drawback that they rely on a fixed image resolution. As the bounding boxes of layout elements naturally vary in their size and aspect ratio, a visual embedding model in the context of \ourwork needs to cope with arbitrary input resolutions. We therefore opted to use the Native-Aspect Flexible Vision Transformer (NaFlexViT), 
which combines aspects of several recent lines of research \cite{tschannen25_naflex, deghani23_navit, beyer23_flexivit}. Based on the cropped image segment of each bounding box, it creates a feature vector of size 768 capturing generic visual information associated with a layout element.

\subsubsection{Textual Embedding.}
Textual features form the second building block in our collection of embeddings. To create them, we first have to retrieve the text from a given document. We opted to use the Tesseract open source OCR engine \cite{smith07_tesseract} as backbone for text retrieval to facilitate reproducibility. Experiments with proprietary OCR engines yielded only marginal performance differences ($<1$\%).
Based on the text contained in a given bounding box, a text embedding model is used to create an abstract representation of the textual content. We again opted to use an open weight model, the multilingual E5 text embedding model \cite{wang24_multilingual_e5} in its small variant, to retrieve a feature vector of size 384.

\subsubsection{Positional Embedding.}
The \ourwork framework extracts positional embeddings directly from annotation data, to capture relative size and position of the layout object, as well as relations to co-occurring layout objects. Specifically, the embedding consists of: (1) the object's own bounding box normalized by the document width and height, (2) the normalized bounding box of its nearest neighbour under Euclidean closest point distance, (3) their distance normalized by the document's diagonal, and (4) the total number of objects within the document, normalized by the dataset-wide maximum. Positional embeddings only rely on the given information of bounding box sizes and positions, making them agnostic to class labels.

\subsection{Layout Object Encoder}
\label{sec:loe}
We propose a novel multimodal encoder architecture, the \gls{loe}, designed to project multiple representations of layout objects in a unified embedding space. By training a model using the \gls{loe} architecture, \ourwork aims to learn a representation integrating visual, textual, and positional information. The model is trained through layout object classification, with the internal representation serving as the layout object embedding.

\begin{figure}[t]
    \centering
    \includegraphics[width=1\linewidth]{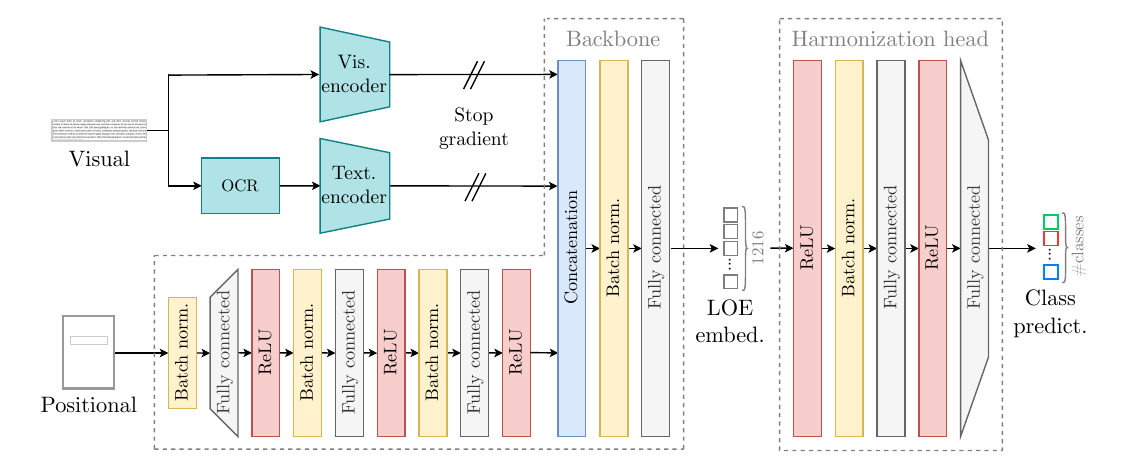}
    \caption{Overview of the \acrlong{loe} architecture. The encoder maps visual, textual, and positional representations of a layout object to a unified embedding. During training, the harmonization head provides supervision through a surrogate classification task.}
    \label{fig:loe_architecture}
\end{figure}

\Cref{fig:loe_architecture} provides an overview of the \gls{loe} architecture. The model processes the positional input embedding vectors through three blocks, each consisting of a batch normalization layer followed by a fully connected layer with $64$ output channels and ReLU activation. The \gls{loe} then concatenates this vector with the visual and textual input representations and processes the result through two analogous blocks that preserve dimensionality. A final fully connected layer produces classification outputs matching the number of classes. The layout object embeddings consist of the features output at the first fully connected layer after concatenation, as illustrated in \cref{fig:loe_architecture}, separating backbone from classification head. Empirical analysis determined this separation as optimal for generating general and transferable layout object representations, distinct from surrogate-specific classification layers.

Given visual, textual, and positional input representations, \ourwork trains the \gls{loe} to learn a weighting of the individual input modalities and integrate them into a unified representation. Through this multimodal feature fusion, the \gls{loe} learns to predict class labels of layout objects given their different input modalities. By training on a large and diverse \gls{dla} dataset, this approach learns a layout object representation generalizing across datasets and class definitions.

Specifically, the \gls{loe} is trained on DocLayNet \cite{pfitzmann22_doclaynet}.
Every epoch, the embedding quality is validated by performing label propagation on the PRImA layout analysis dataset~\cite{antonacopoulos09_prima}. For \ourwork, we use the \gls{loe} checkpoint with the highest pseudo-label accuracy obtained during validation. This selection strategy identifies the checkpoint that generalizes best across datasets and class definitions for predicting pseudo-labels with Label Propagation on unseen datasets.

\subsubsection{Implementation Details.}
The \gls{loe} is trained for $30$ epochs on the DocLayNet training split with a base learning rate of $10^{-4}$, cosine scheduling and cross-entropy loss with label smoothing. Epoch-wise evaluation employs label propagation on $10$ problem instances with $10\%$ re-annotated documents, generated from the PRImA dataset train split. 
All experiments are executed on a single NVIDIA RTX $3090$ GPU.

\subsection{Graph-based Label Propagation} 
Label propagation~\cite{zhu02_labelprop,zhou03_localconsistency} is an algorithm addressing a transductive classification problem. 
It exploits the transductive setting where both labelled and unlabelled instances are available upfront and label predictions are only required for the fixed unlabelled set. Based on the smoothness assumption, Label Propagation leverages the spatial distribution across both labelled and unlabelled data, enabling accurate predictions, even in low supervision settings. Since all bounding box annotated layout objects of a dataset are initially available in the considered problem setting, the algorithm is particularly well-suited for our framework.

Specifically, the algorithm constructs a nearest-neighbour graph where nodes represent samples, each connecting to its $k=20$ nearest neighbours with edge weights reflecting pairwise cosine similarities. In each iteration, the algorithm propagates label information through the graph and fuses it with the initial label information. This process iterates until convergence. Then, the algorithm assigns to each unlabelled sample the label corresponding to the highest class score. Zhou~\etal~\cite{zhou03_localconsistency} derived a closed-form solution for the convergent state, enabling an efficient implementation. For further details, we refer to \cite{zhu02_labelprop,zhou03_localconsistency}.

\section{Experimental Results} \label{sec:experiments}
The evaluation of our \ourwork framework addresses two aspects. It measures firstly the quality of the generated pseudo-labels and secondly, their effectiveness as training data for \gls{dla} models. We assess each aspect separately using two evaluation protocols. Additional experiments investigate the contribution of individual input modalities and compare different \gls{dla} model architectures when training with \ourwork pseudo-labels.

\subsection{Experimental Setup}
Since \ourwork learns a layout object representation generalizing well across datasets, the \gls{loe} trains, validates, and evaluates on distinct datasets. While \ourwork trains on DocLayNet \cite{pfitzmann22_doclaynet} and validates on PRImA, evaluation employs three common \gls{dla} benchmarks: \dfourla \cite{cheng23_d4la}, DocBank \cite{li20_docbank} and PubLayNet \cite{zhong19_publaynet}. To limit computational cost, DocBank and PubLayNet evaluation use $10\%$ subsets, denoted DocBank$_{10\%}$ and PubLayNet$_{10\%}$, respectively. \Cref{tab:dataset_summary} provides an overview of the size and complexity of each dataset.

\begin{table}[t]
    \caption{Summary of experimental datasets showing total number of documents, layout objects, and object categories per dataset. The \textsc{Usage} column indicates whether we use a dataset for training, validation or testing.}
    \centering
    \setlength{\tabcolsep}{6pt}
    \begin{tabular}{l r r r r}
         \toprule
          \textsc{Dataset} & \textsc{Documents} & \textsc{Objects} & \textsc{Classes} & \textsc{Usage} \\
         \midrule
         DocLayNet \cite{pfitzmann22_doclaynet} & \num{80863} & \num{1107470} & $11$ & train\\
         PRImA \cite{antonacopoulos09_prima} & \num{478} & \num{9959} & $6$ & val \\
         \dfourla \cite{cheng23_d4la} & \num{11092} & \num{146846} & $27$ & test \\
         DocBank$_{10\%}$ \cite{li20_docbank} & \num{50000} & \num{671596} & $13$& test \\
         PubLayNet$_{10\%}$ \cite{zhong19_publaynet} & \num{35837} & \num{352289} & $5$& test \\
         \bottomrule
    \end{tabular}
    \label{tab:dataset_summary}
\end{table}

To generate a problem instance consisting of a partially re-annotated dataset, evaluation protocols randomly remove label information from a subset of available documents. The remaining documents represent the manually re-annotated fraction, and the objective is to predict pseudo-labels for the removed annotations. Evaluation assesses pseudo-label accuracy across different fractions of manually re-annotated data, denoted by their re-annotated percentage.

\begin{table}[t]
    \caption{Evaluation results of BBLP default configuration compared to linear classifier baselines using \gls{loe} and visual input features. For each dataset and re-annotation setting, the table reports the mean and standard deviation of pseudo-label accuracies across $10$ random instances in percent.}
    \centering
    \setlength{\tabcolsep}{6pt}
    \begin{tabular}{l r r r r}
         \toprule
         \multirow{2}{*}{\textsc{Dataset}} & \multicolumn{4}{c}{\textsc{Re-annotation fraction}} \\
        
          & $1\%$ & $5\%$ & $10\%$ & $20\%$  \\
         \midrule
         \multicolumn{5}{c}{\textit{Linear classifier baseline (\gls{loe} features)}} \\
         \midrule
         \dfourla & $47.49  \scriptstyle \, \pm 0.94$ & $54.85 \scriptstyle \, \pm 0.53$ & $57.35 \scriptstyle \, \pm 0.58$ & $60.57 \scriptstyle \, \pm 0.68$  \\
         DocBank$_{10\%}$ & $67.24  \scriptstyle \, \pm 3.60$ & $74.66 \scriptstyle \, \pm 0.37$ & $75.13 \scriptstyle \, \pm 0.44$ & $75.55 \scriptstyle \, \pm 0.40$  \\
         PubLayNet$_{10\%}$ & $91.20  \scriptstyle \, \pm 0.36$ & $93.55 \scriptstyle \, \pm 0.09$ & $94.04 \scriptstyle \, \pm 0.04$ & $94.43 \scriptstyle \, \pm 0.03$  \\
         \midrule
         \multicolumn{5}{c}{\textit{Linear classifier baseline (visual features)}} \\
         \midrule
         \dfourla & $42.40  \scriptstyle \, \pm 0.94$ & $47.52 \scriptstyle \, \pm 0.99$ & $48.44 \scriptstyle \, \pm 1.54$ & $49.32 \scriptstyle \, \pm 0.86$  \\
         DocBank$_{10\%}$ & $67.11  \scriptstyle \, \pm 3.44$ & $70.80 \scriptstyle \, \pm 1.17$ & $70.02 \scriptstyle \, \pm 4.54$ & $72.88 \scriptstyle \, \pm 3.22$  \\
         PubLayNet$_{10\%}$ & $92.92  \scriptstyle \, \pm 0.15$ & $93.69 \scriptstyle \, \pm 0.07$ & $93.95 \scriptstyle \, \pm 0.05$ & $94.21 \scriptstyle \, \pm 0.06$  \\
         \midrule
         \multicolumn{5}{c}{\textit{\ourwork (ours)}} \\
         \midrule
         \dfourla & $57.32  \scriptstyle \, \pm 0.58$ & $64.71 \scriptstyle \, \pm 0.38$ & $67.65 \scriptstyle \, \pm 0.27$ & $69.42 \scriptstyle \, \pm 0.21$  \\
         DocBank$_{10\%}$ & $78.90 \scriptstyle \, \pm 0.18$ & $81.55 \scriptstyle \, \pm 0.07$ & $82.08 \scriptstyle \, \pm 0.07$ & $82.32 \scriptstyle \, \pm 0.05$  \\
         PubLayNet$_{10\%}$ & $93.95  \scriptstyle \, \pm 0.11$ & $94.75 \scriptstyle \, \pm 0.05$ & $94.93 \scriptstyle \, \pm 0.04$ & $95.02 \scriptstyle \, \pm 0.04$  \\
         \bottomrule
    \end{tabular}
    \label{tab:pseudo_label_accuracy}
\end{table}

\subsection{Pseudo-Label Quality}
\label{sec:pseudo-label quality}
To assess pseudo-label quality, we evaluate \ourwork on the train split of each evaluation dataset under the re-annotated fractions of $1$, $5$, $10$, and $20$ percent. For each setting, we sample $10$ problem instances and measure accuracy on the unlabelled fraction of the dataset.

To the best of our knowledge, no prior work in the \gls{dla} domain addresses the re-annotation of bounding boxes. Therefore, we cannot compare our approach to existing methods. To establish a baseline, we compare Label Propagation to a simple linear classifier with softmax activation, as a method for generating pseudo-labels from layout object representations. Two variants of the linear classifier are tested. The first variant uses our \gls{loe} features, while the second uses visual input features only. Training is started from scratch on the re-annotated fraction of each problem instance for $100$ epochs with a learning rate of $0.1$. The trained classifier then predicts bounding box pseudo-labels for the unlabelled documents.

\Cref{tab:pseudo_label_accuracy} reports the results for the pseudo-label quality evaluation protocol.
Across all datasets and approaches, pseudo-label accuracy increases with additional supervision, with diminishing gain at higher supervision settings. Results are strongest for PubLayNet$_{10\%}$, while \dfourla exhibits the lowest average accuracy scores, reflecting the respective number of classes. \ourwork achieves the highest average accuracies across all datasets and supervision settings, exceeding $90\%$ on PubLayNet$_{10\%}$, around $80\%$ on DocBank$_{10\%}$ and below $70\%$ on \dfourla. Both linear classifier baselines reach considerably lower scores. The margins are most pronounced for \dfourla, where \ourwork outperforms the \gls{loe} features linear classifier by roughly $10$ percentage points and the visual features baseline by more than $15$ percentage points across all supervision settings. On DocBank$_{10\%}$ margins remain smaller yet significant, with \ourwork surpassing the \gls{loe} features and visual features baselines in all settings by $6$ and $9$ percentage points, respectively. Improvements on PubLayNet$_{10\%}$ are notably smaller, as both baselines consistently achieve average accuracies over $91\%$, yet \ourwork yields a consistent advantage.

\subsection{Document Layout Analysis with BBLP}
\label{sec:dla_evaluation}
Object detection models have demonstrated robustness against label-noise to a certain extent~\cite{hadsell06_contrastive}. However, training with \ourwork pseudo-labels only remains practical if \gls{dla} models can tolerate the inherent amount of label noise. To further investigate the \ourwork pseudo-label quality and to examine the effect of the inherent label noise on the downstream task of \gls{dla}, we train an object detector on different label configurations. This evaluation compares the generated pseudo-labels against a gold standard and a lower performance bound baseline as \gls{dla} training data. 

For the train split of each evaluation dataset, we randomly sample a single problem instance using $10\%$ re-labelled data, generate pseudo-labels with \ourwork for the unlabelled fraction, and train on the complete dataset, including manually and pseudo-labelled documents. The gold standard baseline model trains on the fully ground-truth annotated dataset, while the lower baseline uses only the $10\%$ ground-truth data, representing the manually re-annotated documents.

After each epoch, we evaluate the current model checkpoint on the corresponding dataset test split and report mAP across all training epochs.

All models trained in this experiment follow the DINO architecture with a ResNet-50 backbone and 4-scale feature maps \cite{zhang23_dino}. In each configuration, the model is initialized with weights pre-trained on the COCO dataset~\cite{lin14_COCO} and is trained for $18$ epochs on the corresponding evaluation dataset. The remaining hyperparameters follow those in~\cite{zhang23_dino}. 

\begin{figure}[t]
    \centering
    \includegraphics[width=1\linewidth]{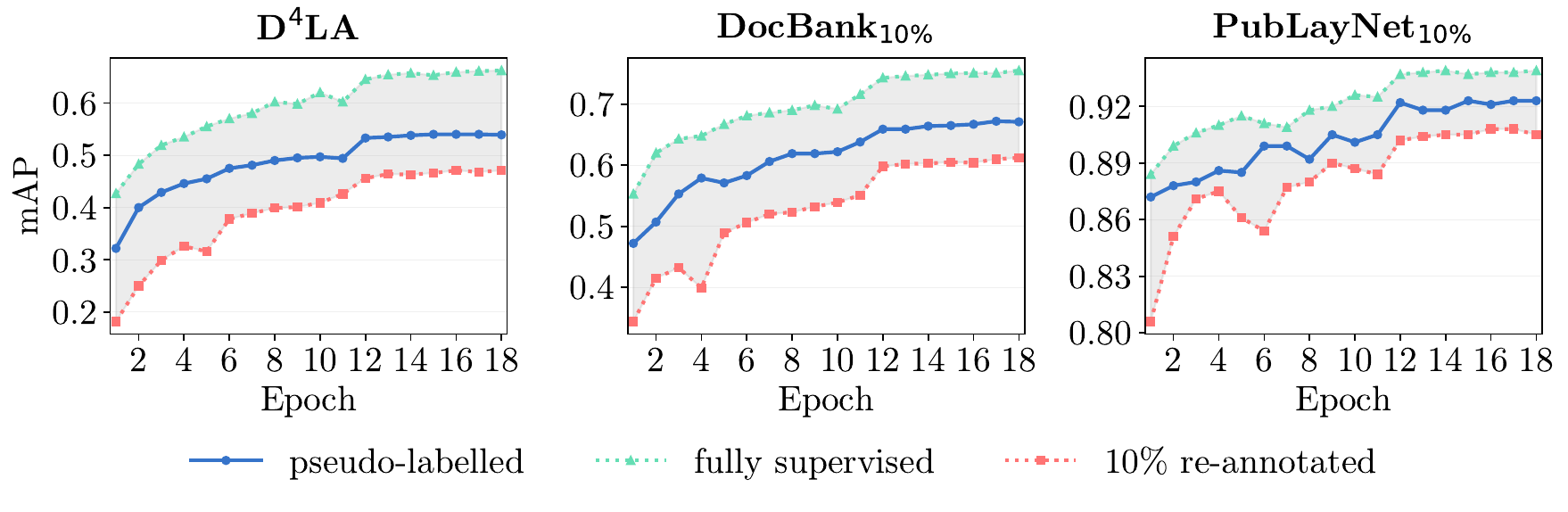}
    \caption{Training performance of a DINO object detector trained on \ourwork pseudo-labels. For each evaluation dataset, curves show mAP scores on the test split during training for models trained on pseudo-labelled, $10\%$, and $100\%$ ground-truth annotated documents.}
    \label{fig:od_evaluation_mAP_curve}
\end{figure}

\Cref{fig:od_evaluation_mAP_curve} visualizes the mAP progression on the test splits during training for the different label configuration and evaluation datasets. Across all evaluation datasets and epochs, the pseudo-label trained model consistently exceeds the lower baseline trained on $10\%$ manually re-annotated documents, while remaining below gold standard trained on the full ground-truth annotated dataset. On \dfourla, \ourwork reaches a converged mAP of $53.9\%$, compared to $47.1\%$ and $66.0\%$ for the lower and gold standard baseline, respectively. Similarly, on DocBank$_{10\%}$ the \ourwork pseudo-labels trained model achieves finally $67.1\%$ mAP, surpassing the lower baseline by $5.9$ percentage points and falling $8.4$ percentage points short to the gold standard baseline. On PubLayNet$_{10\%}$, performance of all three models is comparable. The pseudo-label trained model reaches a mAP score of $92.3\%$ at convergence, which is only $1.8$ percentage points above the lower and $1.6$ percentage points below gold standard baseline.

\subsection{Additional Experiments}

To better understand the contribution of each modality and to evaluate an alternative \gls{dla} model trained on the generated pseudo-labels, we conduct two additional experiments. 

\subsubsection{Influence of Individual Modalities.} As outlined in \cref{sec:loe}, \ourwork multimodal feature encoding integrates visual, textual, and positional information in a joint embedding. We hypothesize that training on a large and diverse dataset enables the \gls{loe} to appropriately weight the influence of individual input modalities, yielding a layout object representation superior to any individual input representation. To investigate this hypothesis, this ablation compares the default \gls{loe} representation against visual and textual input modality representations, as well as against \gls{loe} variants integrating only one or two modalities.

These \gls{loe} variants concatenate only the remaining features in the concatenation layer. The resulting feature vector size determines the dimensionality of all subsequent harmonization layers, reducing the overall parameter amount compared to the standard \gls{loe}. Training follows the procedure outlined in \cref{sec:loe}. For evaluating each representation setting, this ablation adopts the pseudo-label quality evaluation protocol outlined in \cref{sec:pseudo-label quality} and reports results only for the $10\%$ re-labelled setting.

\begin{table}[t]
    \caption{Impact of individual modalities on \ourwork pseudo-label accuracy. We report average accuracies and standard deviations with 10\% re-labelled data, comparing raw visual and textual input features against features from \gls{loe} variants trained on combinations of modalities. Bold values indicate the best and underlined values the second-best result per dataset.}
    \centering
    \setlength{\tabcolsep}{6pt}
    \begin{tabular}{l l l r r r}
      \toprule
      \multicolumn{3}{c}{\textsc{Modalities}} & \multicolumn{3}{c}{\textsc{Dataset}} \\
      vis. & text. & pos. &\dfourla & DocBank$_{10\%}$ & PubLayNet$_{10\%}$ \\

      \midrule 
      \multicolumn{6}{c}{\textit{Input embeddings}} \\
      \midrule
      \checkmark & & & $56.64 \scriptstyle \, \pm 0.26$ & $81.53 \scriptstyle \, \pm 0.07$ & $\textBF{95.07} \scriptstyle \, \pm 0.05$ \\
       & \checkmark &  & $62.86 \scriptstyle \, \pm 0.32$ & $73.47 \scriptstyle \, \pm 0.16$ & $84.81 \scriptstyle \, \pm 0.10$ \\  
      
      \midrule
      \multicolumn{6}{c}{\textit{\gls{loe} embeddings}} \\
      \midrule

       \checkmark &  &  & $55.61 \scriptstyle \, \pm 0.28$ & $78.73 \scriptstyle \, \pm 0.06$ & $94.39 \scriptstyle \, \pm 0.03$ \\
       & \checkmark &  & $61.14 \scriptstyle \, \pm 0.33$ & $71.87 \scriptstyle \, \pm 0.10$ & $83.81 \scriptstyle \, \pm 0.06$ \\
       &  & \checkmark & $38.25 \scriptstyle \, \pm 0.12$ & $63.04 \scriptstyle \, \pm 0.06$ & $80.97 \scriptstyle \, \pm 0.05$ \\  

       & \checkmark & \checkmark & $65.02 \scriptstyle \, \pm 0.29$ & $75.56 \scriptstyle \, \pm 0.15$ & $88.15 \scriptstyle \, \pm 0.12$ \\
      \checkmark &  & \checkmark & $57.52 \scriptstyle \, \pm 0.23$ & $79.19 \scriptstyle \, \pm 0.06$ & $94.45 \scriptstyle \, \pm 0.03$ \\
      \checkmark & \checkmark &  & $\underline{66.81} \scriptstyle \, \pm 0.24$ & $\underline{81.76} \scriptstyle \, \pm 0.07$ & $94.85 \scriptstyle \, \pm 0.04$ \\

       \checkmark & \checkmark & \checkmark & $\textBF{67.65} \scriptstyle \, \pm 0.27$ & $\textBF{82.32} \scriptstyle \, \pm 0.05$ & $\underline{94.93} \scriptstyle \, \pm 0.04$ \\
     \bottomrule
    \end{tabular}
    \label{tab:modality ablation}
\end{table}

\Cref{tab:modality ablation} reports the pseudo-label accuracies for all modality settings. On \dfourla and DocBank$_{10\%}$, the original \ourwork configuration using all three input modalities reaches the highest pseudo-label accuracies, while on PubLayNet$_{10\%}$ it can compete with the visual input embedding. Among single-modality \gls{loe} variants, the text modality contributes most to performance on \dfourla, whereas the visual modality achieves the highest individual scores on DocBank$_{10\%}$ and PubLayNet$_{10\%}$. The positional embedding achieves the lowest scores across all datasets. Each multi-modality configuration consistently outperforms the corresponding single-modality configurations, confirming that the \gls{loe} leverages complementary information across modalities.

\subsubsection{Alternative \acrshort{dla} Model Architecture.}
The experiment in \cref{sec:dla_evaluation} demonstrates that \gls{dla} models can generally exploit additional supervision provided by \ourwork pseudo-labels, despite the inherent amount of label noise. However, the evaluation considers only one specific model architecture. This experiment evaluates an alternative \gls{dla} model architecture trained on the generated pseudo-labels, to verify that noise tolerance generalizes across model architectures, rather than being specific to DINO model architecture.

Specifically, this experiment trains a two-stage Faster R-CNN \cite{ren15_fasterrcnn} with a ResNet-$50$ backbone for $18$ epochs. This experiment focuses exclusively on \dfourla, as the dataset has the highest pseudo-label noise ratio and label noise effects on training are potentially most pronounced. Analogous to \cref{sec:dla_evaluation}, training uses the same set of generated pseudo-labels and compares the model to a gold standard and a $10\%$ manually re-annotated baseline.

\Cref{tab:alt_dla_arch_ablation} reports the final mAP scores for all label configurations and both model architectures. Similar to the DINO architecture, the Faster R-CNN pseudo-label configuration consistently outperforms the lower baseline by a significant margin but performs below the gold standard, though the relative margin to the lower baseline is smaller compared to DINO. Furthermore, Faster-CNN converges at notably lower mAP values than DINO across all label configurations, which can be attributed to DINO's inherent performance advantage over Faster R-CNN, as documented in \cite{zhang23_dino}.

\begin{table}[t]
    \caption{Performance comparison across \gls{dla} architectures. The table reports final mAP scores in percent on \dfourla after training under three label configurations: \ourwork pseudo-labels, fully supervised gold standard baseline and $10\%$ manually re-annotated baseline.}
    \centering
    \setlength{\tabcolsep}{6pt}
    \begin{tabular}{l r r r}
      \toprule
      \textsc{DLA Model} & \textsc{Pseudo-lab.} & \textsc{Fully sup.} & \textsc{$10\%$ re-ann.} \\
      \midrule
      DINO & 53.9 & 66.2 & 47.1 \\
      Faster R-CNN & 42.0 & 54.2 & 37.9 \\
      
     \bottomrule
    \end{tabular}
    \label{tab:alt_dla_arch_ablation}
\end{table}

\section{Discussion}\label{sec:discussion}
\Cref{sec:pseudo-label quality} demonstrates that \ourwork can reliably assign correct labels to the majority of layout objects, even under minimal supervision and for datasets with numerous classes such as \dfourla. A higher fraction of manually re-annotated documents leads to improved pseudo-label accuracy on the remaining documents. This gain is most pronounced under low supervision, as additional labelled samples become increasingly redundant and contribute less novel information with growing supervision. Furthermore, the marginal accuracy gains from additional supervision correlate with the number of classes in the dataset. Such behaviour aligns with the intuition that datasets with many classes require more labelled instances to represent each class adequately. Standard deviations remain negligibly low for all datasets and re-labelled fractions, indicating minimal dependency on the choice of documents for manual re-annotation. 

The performance advantage of \ourwork compared to the linear classifier baselines likely stems from the transductive nature of the Label Propagation algorithm. While linear classifiers rely exclusively on labelled instances as training data for generating pseudo-labels, Label Propagation additionally leverages the spatial distribution of unlabelled data to produce label predictions. Moreover, the \gls{loe} feature baseline outperforms the visual feature baseline on \dfourla and DocBank$_{10\%}$, but shows comparable performance on PubLayNet$_{10\%}$. This trend likely reflects dataset complexity: PubLayNet$_{10\%}$ contains only five visually distinguishable classes, while textual and positional cues integrated in the \gls{loe} features prove advantageous for more complex class structures.

Furthermore, experiments demonstrate that \ourwork pseudo-labels are well-suited for \gls{dla} model training. Even under high noise ratios of almost $30\%$ for the \dfourla dataset, the models leverage the additional supervision and outperform the baseline trained on only manually re-annotated documents by a significant margin. In comparison to the gold standard baseline, the pseudo-label trained model's relative mAP surpasses the label accuracy of its training data. This finding supports the hypothesis that \gls{dla} models tolerate label noise to a certain extent and exploit incorrectly labelled bounding boxes, as positional information remains correct and semantic similarity often exists between predicted and ground-truth labels. These trends persist across \gls{dla} model architectures in our experiments, indicating general applicability.

As the results of the ablation study demonstrate, the quality of \gls{loe} features consistently improves when using more modalities. During \gls{loe} training, the encoder learns to effectively integrate multi-modal input features in a joint general embedding through appropriate weighting. Across modalities, the textual features have the greatest impact on Label Propagation performance for \dfourla, while visual embeddings contribute most for DocBank$_{10\%}$ and PubLayNet$_{10\%}$. A likely explanation lies again in the different class definition complexities. \dfourla contains $27$ classes, many of which are text-based with similar visual representations. Conversely, the five classes of PubLayNet$_{10\%}$ are plausibly distinguishable by visual features alone. The positional modality achieves the lowest individual performance across datasets, yet consistently improves representation quality when combined with other modalities. Evidently, positional information provides valuable complementary cues, despite being insufficient for classification alone.

\subsubsection{Limitations.}
While we demonstrate the potential of Label Propagation for semi-automatic \gls{dla} dataset re-annotation, several limitations need to be acknowledged. The \ourwork framework incorporates spatial information from the annotation data, but does not consider class labels prior re-annotation. Given that such refinements typically follow hierarchical patterns, original class labels often hint at semantic relationships and can provide valuable cues to predict refined class labels. Furthermore, this work assesses the quality of pseudo-labels only by label accuracy across the entire dataset. However, layout object classes are naturally highly imbalanced. Per-class evaluation metrics could reveal insights in the structure of label noise, such as common class confusions.

\section{Conclusions and Future Work} \label{sec:conclusions}
This work addresses the challenge of reducing manual re-annotation effort for taxonomy adaptation of \gls{dla} datasets. Given a small subset of manually re-annotated documents, the proposed \ourwork framework generates pseudo-labels for layout objects in the remainder of the dataset. The novel \gls{loe} model architecture integrates visual, textual, and positional features into unified layout object representations. \ourwork leverages these representations to transfer label information from labelled to unlabelled object instances in the dataset.

Experiments demonstrate that all three modalities provide complementary information, which the \gls{loe} effectively integrates into a higher-performing joint representation. Moreover, Label Propagation proves particularly effective for the semi-automatic dataset re-annotation, since the algorithm leverages spacial structure of instance representations across both labelled and unlabelled data. Despite the inherent amount of label noise, \gls{dla} models can take advantage of the additional supervision provided by
\ourwork pseudo-labels, improving performance significantly compared to training on manually re-annotated documents alone.

The identified limitations provide directions for future research. One promising extension would integrate prior class label information in the \ourwork framework. For instance, the Label Propagation $k$-nearest-neighbour graph could enhance edge weights based on original class agreement, facilitating label information flow between semantically related objects. Additionally, we plan to investigate our approach in conjunction with established multimodal encoders, such as LayoutLM \cite{huang22_layoutlmv3}. Both directions offer potential for substantially reducing manual re-annotation effort when refining class definitions of \gls{dla} datasets.


%
%


%
%

\bibliographystyle{splncs04}
\bibliography{bib}

\end{document}